\documentclass[]{fairmeta}
\usepackage{xcolor}

\usepackage{hyperref}
\usepackage{url}
            
\usepackage[utf8]{inputenc} 
\usepackage[T1]{fontenc}    
\usepackage{url}            
\usepackage{booktabs}       
\usepackage{amsfonts}       
\usepackage{nicefrac}       
\usepackage{enumitem}
\usepackage{subcaption}
\usepackage{wasysym}
\usepackage{textgreek}
\usepackage[table,xcdraw]{xcolor}
\usepackage{mdframed}

\usepackage{microtype}      
\usepackage{graphicx}
\usepackage{adjustbox}
\usepackage{float}               

\usepackage{amsmath,amssymb,amsthm,bm,dsfont}
\usepackage{array, booktabs, makecell}
\usepackage{multirow}
\usepackage{cleveref}

\usepackage{pifont}

\usepackage[T1]{fontenc} 
\usepackage{lmodern}
\usepackage{courier}    
\usepackage{relsize}


\title{Accurate Decoding of Natural Sentences from Non-Invasive Brain Recordings}

\author[2,3*]{Mingfang (Lucy) Zhang}
\author[1,7*]{Jarod Lévy}
\author[1]{Cedric Rommel}
\author[1]{Jérémy Rapin}
\author[2,8]{Corentin Bel}
\author[8]{Julie Bonnaire}
\author[5]{Daniel Nieto}
\author[3,6\dagger]{Pierre Bourdillon}
\author[4,5\dagger]{Svetlana Pinet}
\author[1\dagger]{Stéphane d'Ascoli}
\author[7\dagger]{Thomas Moreau}
\author[1\dagger]{Jean-R\'emi King}

\affiliation[1]{Meta AI}
\affiliation[2]{École Normale Supérieure, Université PSL, CNRS}
\affiliation[3]{Hospital Foundation Adolphe de Rothschild}
\affiliation[4]{Univ. Lille, CNRS, UMR 9193-SCALab-Sciences Cognitives et Sciences Affectives, F-59000 Lille, France}
\affiliation[5]{Basque Center on Cognition, Brain and Language, San Sebastian}
\affiliation[6]{Paris Cité University}
\affiliation[7]{Inria, Université Paris-Saclay, Palaiseau, France}
\affiliation[8]{CNRS, INSERM, CEA, Neurospin center, Gif-sur-Yvette}

\contribution[*]{equal contribution}
\contribution[\dagger]{These authors jointly supervised this work}

\date{\today}
\metadata[Code]{\url{https://github.com/facebookresearch/brain2qwerty}}
\correspondence{
\email{jarod@meta.com},
\email{lucy.zhang@psl.eu},
\email{jeanremi@meta.com}
}

\begin{document}

\abstract{

Restoring communication for people who have lost the ability to speak or move after a brain injury is a major challenge.
While intracranial implants now enable high-performing brain-computer-interfaces, non-invasive alternatives are still lagging behind.
Here, we present Brain2Qwerty v2, a model that can decode the production of natural sentences solely from real-time magnetoencephalography (MEG) recordings. By collecting 22,000 sentences typed by nine subjects, each recorded for 10 hours, our model leverages character, word and sentence-level representations to achieve an average word error rate (WER) of 39\%. For our best participant, the model accurately decodes half of the sentences with one word error or less. Critically, decoding accuracy log-linearly improves with data volume, suggesting that the performance gap with intracranial approaches could be partially bridged through data scaling. 
We show that AI enables this performance in three main ways: 
the substitution of hand-crafted pipelines for event detection with deep learning, 
the finetuning of large language models to extract semantic representations, 
and the deployment of AI agents to iteratively refine our decoding pipeline via automated code development.
Together, these results show that non-invasive brain-to-text decoding starts to operate at a level of accuracy previously thought exclusive to surgical implants, opening a path toward safe and efficient brain-computer-interfaces.
}

\maketitle

\section{Introduction}

\begin{figure}[!t]
    \centering
    \includegraphics[width=\linewidth, trim={0 0.9cm 0 0}, clip]{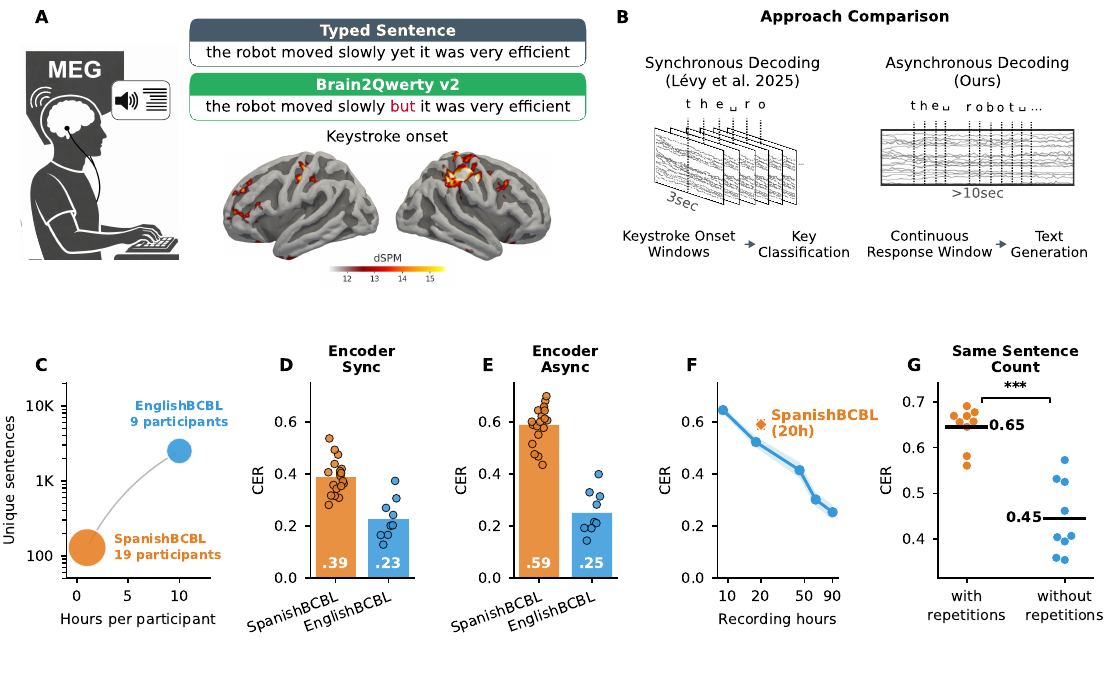}
    \caption{\textbf{Asynchronous MEG decoding is unlocked by recording scale and variety}. \\
    \textbf{ A.} Experimental protocol. Left. We recorded healthy volunteers for 10 hours each using Magnetoencephalography (MEG) while they typed natural sentences they heard a few seconds prior. Right. Average MEG source reconstruction at the time of key press suggest that MEG primarily picks neural activity in the motor cortex. \textbf{B.} Approaches for brain-to-text decoding. Synchronous decoding consists in classifying the character from windows time-locked to each keystroke (e.g. \citet{levy2025brain}). Asynchronous decoding consists in decoding text from a continuous brain signal, and can thus be applied in real-time, although with some potential delays (e.g. \citet{feghhi2025time}). \textbf{C.} Quantity (hours per participant) and diversity (number of unique sentences) of our dataset (EnglishBCBL) as compared to \citet{levy2025brain} (SpanishBCBL). 
    \textbf{D.}~Character-error-rate (CER) for the synchronous encoder of \citet{levy2025brain}. Each coloured dot is one subject; the bar plots the across-subject mean.
    \textbf{E.}~Same as D for our asynchronous encoder. \textbf{F.}~Scaling of the asynchronous encoder CER as a function of the amount of training data (log scale), expressed as total recording hours per subject (test set fixed across all points). The blue curve is the across-subject mean on our EnglishBCBL dataset. The orange diamond places SpanishBCBL on the same axis after training our asynchronous encoder on it.
    \textbf{G.}~Impact of sentence-list variety on asynchronous encoder CER at matched total sentence count: 128 unique sentences $\times$ 2 repetitions (SpanishBCBL protocol, orange) versus 256 unique sentences (EnglishBCBL protocol, blue). Two-sided Mann-Whitney $U$ test across the $n{=}9$ subjects per condition: (***) denotes p < 0.001.
    Across all panels, CER is computed per sentence, then averaged within each subject, and finally averaged across subjects.}
    \label{fig:fig1}
\end{figure}

Invasive brain-computer interfaces (BCIs) have recently restored communication in individuals, who, after a brain lesion or disorder, have lost the ability to speak or move. Several patients suffering from anarthria, locked-in syndrome, or amyotrophic lateral sclerosis (ALS) and implanted with electrodes over the motor cortex have recently been able to produce language via attempted speech~\citep{moses2021neuroprosthesis,willett2023neuroprosthesis, card2024accurate, wairagkar2024voice}, handwriting~\citep{willett2021handwriting}, and typing~\citep{pandarinath2017high,judeRestoringRapidNatural2026}, with decoding speed and accuracy that approach those of natural speech~\citep{judeRestoringRapidNatural2026}. Despite these breakthroughs, the invasive nature of intracranial implants is a challenge for large-scale clinical translation. Neurosurgery implies significant medical risks and good neural recording can be challenging to maintain over long time-periods because of neuroinflammatory responses. Providing specialized surgical care to a broad patient population would also pose considerable logistic challenges. 


As an alternative, many non-invasive BCIs have been proposed over the years~\citep{Mak2009, abiri2019review, chevallier2024, bodien2024cognitive}. However, each of these proposal faces important challenges. The poor \emph{signal-to-noise ratio} of electroencephalography (EEG) has limited BCIs to complex and highly-demanding tasks, making them effectively impractical for patients \citep{Mak2009,lotte2018review}. Similarly, the limited \emph{temporal resolution} of functional Magnetic Resonance Imaging intrinsically limits its use for communication~\citep{owen2006detecting,tang2023semantic}. Magnetoencephalography (MEG) may provide a promising alternative~\citep{Baillet2017,levy2025brain, dascoli2025towards,landau20252025}, but still lags behind the performance of invasive recordings. In particular, \citet{levy2025brain} recently decoded typed text with a 32\% character error rate (CER) by training a classifier time-locked to each key stroke. However, this approach faces three main challenges. First, it requires knowing the timing of individual keystrokes, and hence limits real-time usage. Second, keystroke classification does not guarantee effective sentence reconstruction, as misclassified characters lead to incomprehensible output. 
Third, this study is based on a relatively small amount of data: while invasive BCIs typically make use of thousands of sentences produced over 10-40 hours~\citep{card2024accurate, willett2023neuroprosthesis}, \citet{levy2025brain} only had participants recorded for 1 hour each. 

To address these three challenges, we introduce Brain2Qwerty v2, a deep learning framework designed to decode a 22,000-sentence corpus from non-invasive magnetoencephalographic (MEG) recordings. We trained the model on a delayed-typing task performed by nine healthy volunteers across 90 total recording sessions. In each trial, participants listened to a sentence via headphones, waited through a forced delay, and then typed the corresponding text. Our decoding study focuses on the neural activity during the language production phase. This new dataset presents 10 times more data per subject than~\citet{levy2025brain} and spans a much larger diversity of sentences (Figure \ref{fig:fig1}C). 
Building on this dataset, our approach leverages AI in three main ways: First, we resolve the real-time challenge with a Connectionist Temporal Classification (CTC) objective~\citep{graves2006connectionist}. Second, we design the model to generate meaningful sentences by fine-tuning a Large Language Model (LLM). Third, we employ `auto-research' AI agents \citep{zhao2025automated, starace2025paperbench} to autonomously write the code that optimizes our decoding pipeline.

\section{Results}



\subsection{Decoding individual keystrokes}
\paragraph{Source reconstruction confirms expected networks activation during typing.}
To inspect subjects' brain activity, we first perform a source reconstruction analysis at keystroke onset. The results reveal a bilateral activation of the primary motor cortex (M1) and supplementary motor area (SMA). Notably, the activation is more spatially extended in the right hemisphere (Figure \ref{fig:fig1}A), a pattern consistent with the contralateral organization of M1 in a right-handed cohort and with the established role of SMA in bimanual motor coordination~\citep{Swinnen2004}. Additional residual activity surrounding keystrokes can be detected in the left dorso-lateral and infero-frontal gyri and temporo-parietal junction (Figure \ref{fig:source_times}), consistent with the language network \citep{ojemann1991cortical,fedorenko2024language}.

\paragraph{Increased dataset size unlocks competitive asynchronous decoding.}
Decoding keystrokes is a good starting point for decoding typed sentences. 
To remove the constraint of needing keystroke timings in the synchronous approach in prior work \citep{levy2025brain}, here, we implement \textit{asynchronous} decoding, which outputs a sequence of predictions from a continuous window of response (Figure \ref{fig:fig1}B).
To this end, we adopt the Connectionist Temporal Classification (CTC) objective \citep{graves2006connectionist}, which learns to align variable-length MEG recordings to keystroke sequences without requiring onset timing of each keystroke. Such asynchronous decoding with CTC, however, is a substantially harder problem. Therefore, we first ask whether the scale and diversity of our new typing MEG dataset are sufficient to close the gap with synchronous performance, bringing us closer to decoding full sentences without needing keystroke timings.

To quantify the benefit of data scale, we compare the CER on the predictions of the synchronous encoder (Encoder Sync) used in \cite{levy2025brain} and the asynchronous encoder (Encoder Async) used in this work on both the low-data SpanishBCBL dataset~\citep{levy2025brain} and our new, larger, EnglishBCBL dataset (Figure \ref{fig:fig1}D, E). Both encoders comprise a BrainModule \citep{defossez2023decoding}, followed by a Transformer for grouping learned keystroke representations (Encoder Sync) or a Conformer \citep{gulati2020conformer} operating over full time-frame representations (Encoder Async). With Encoder Sync, scaling from SpanishBCBL to EnglishBCBL substantially improved synchronous decoding performance (CER: $0.39\pm 0.02$ → $0.23\pm 0.03$). Critically, the benefit of scale was even more pronounced for the Encoder Async (CER: $0.59\pm 0.02$→ $0.25\pm 0.03$), which nearly matched synchronous performance on EnglishBCBL. Although Encoder Sync outperformed Encoder Async on the low-data SpanishBCBL dataset, this gap collapsed to just 2\% on EnglishBCBL. Overall, these results confirm that the scale of the dataset is a decisive factor enabling asynchronous decoding to approach the performance ceiling set by synchronous methods.

\paragraph{Asynchronous performance scales log-linearly with data quantity.} 
Next, we examine how the performance of the asynchronous encoder scales with training data. We retrain a series of the Encoder Async models on progressively larger subsets of the EnglishBCBL dataset (Figure \ref{fig:fig1}F). CER decreases consistently with recording hours and shows no sign of saturation at the current data ceiling of approximately 90 hours pooled across participants. The scaling follows a clear log-linear trend (Pearson $r = -0.99$, $p = 1.1 \times 10^{-3}$, $R^2 = 0.98$ between $\log_{10}$(hours) and the across-subject mean CER over the 5 training-fraction conditions; slope $= -0.39$ CER per decade), suggesting that further performance gains are achievable with additional data collection.


\paragraph{Sentence diversity independently improves decoding performance.}
Interestingly, on the SpanishBCBL dataset Encoder Async's CER is $0.59 \pm 0.02$ (Figure \ref{fig:fig1}F), significantly higher than training on a similar amount of data from our EnglishBCBL dataset ($0.52 \pm 0.02, p<0.05$). Could this performance gap be due to the difference in language variety across the two datasets? In the SpanishBCBL dataset, each unique sentence was repeated twice by each participant, while in the EnglishBCBL dataset, each unique sentence was only typed once by each participant. In addition, the SpanishBCBL dataset had a small variety of part-of-speech and syntactic structures. To isolate the effect of language variety from quantity, we constructed two controlled datasets matched for total sentence count and number of subjects ($n=9$), but differing in the number of unique sentences: one drawn from SpanishBCBL (128 unique sentences × 2 repetitions per participant) and one from EnglishBCBL (256 unique sentences, each typed once). Encoders trained on non-repeated sentences achieved significantly lower CER than those trained on repeated sentences (CER: $0.45\pm 0.03$ vs. $0.65\pm 0.01$; $p<0.001$; Figure~\ref{fig:fig1}G), demonstrating that sentence diversity constitutes an independent axis of data quality, distinct from total sample volume. Together, these findings show that both the quantity and the variety of training sentences are critical determinants of encoder performance in non-invasive brain-to-text decoding, even without any language models.


\subsection{Decoding natural sentences from MEG with Brain2Qwerty v2}

\paragraph{Brain2Qwerty v2 leverages a multi-level architecture to extract semantic representations.}
Approaches using keystroke decoding focus on extracting signal representations that capture information at the character level. While this step is critical for decoding the typed text, it does not leverage the statistical regularity of natural language, such as that encoded in Large Language Models (LLMs). The above asynchronous keystroke decoder is neither designed to (1) extract semantic representations nor to (2) efficiently leverage the statistical regularities of natural language. 



\begin{figure}[t]
    \centering
    \includegraphics[width=\linewidth]{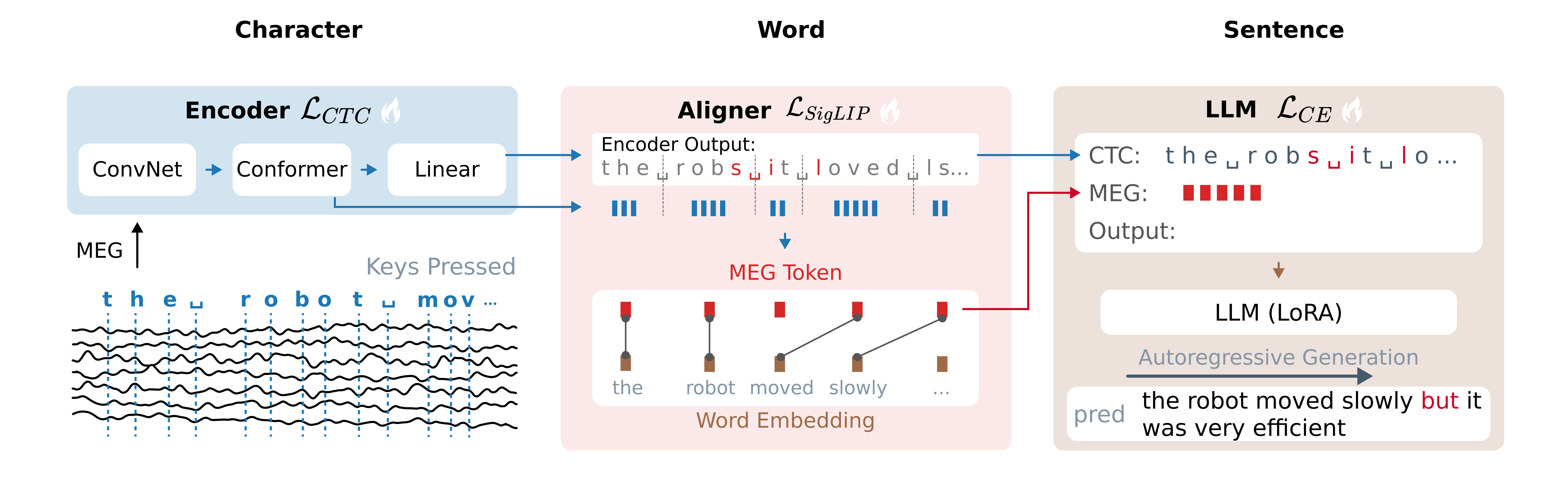}
    \caption{\textbf{Brain2Qwerty v2 architecture}.\\
    Our pipeline is solely input with the continuous MEG recording corresponding to an entire typed sentence and outputs the decoded sentence thanks to three jointly-optimized modules.  
    First, the Encoder is trained with a CTC loss \citep{graves2006connectionist} to extract character-level representations from brain activity, and outputs both MEG Embeddings and a sequence of characters. Second, the Aligner learns, with a SigLIP loss \citep{zhai2023sigmoid}, to group and align the MEG embeddings with the true Word Embeddings. Finally, a Large Language Model (LLM) is input with both the MEG tokens and the Encoder's text to generate the correct sentence autoregressively and with LoRA finetuning \citep{huLoRALowRankAdaptation2021}. 
    }
    \label{fig:arch}
\end{figure}

To utilize higher-level semantic information in the neural recordings and the power of LLMs, we developed a three-level architecture (Figure~\ref{fig:arch}) designed to jointly extract character, word, and sentence-level information: 

\begin{itemize}
    \item The Encoder Async, hereafter referred to as Encoder, is trained with a CTC objective \citep{graves2006connectionist} to output a sequence of keystroke predictions. 
    \item The Aligner connects the neural data to words. By using a contrastive learning method (SigLIP loss; \citealt{zhai2023sigmoid,dascoli2025towards}), it trains the model to directly align the processed brain signals with their correct word representations.
    \item The LLM part finetunes an LLM with Low Rank Adapters (LoRA;~\citealt{huLoRALowRankAdaptation2021}) and a cross-entropy loss to autoregressively generate the target sentence when prompted with both the CTC keystroke predictions and the neuro tokens. 
\end{itemize}

A fundamental challenge in this is how to bridge the gap between continuous MEG embeddings from the Encoder to the word-level tokens expected by LLMs. To address this we design a CTC tokenizer which chunks continuous MEG embeddings based on where the `space' key is in the keystroke predictions. Embeddings between space predictions are designated as MEG word-chunk, which is then averaged and mapped into the LLM's word-embedding space by a small MLP. We train the MLP with a word-level contrastive loss and dynamic time warping (DTW), which pairs each projected MEG token with the closest target word during training~\citep{sakoe1978dynamic}. 

With this three part pipeline and a tokenizer that creates discrete word-like tokens from continuous MEG embeddings, our Brain2Qwerty v2 pipeline can utilize both the keystroke-, word- and contextual-representations to decode the sentence. We train our pipeline on the EnglishBCBL dataset. 




\begin{figure}[t!]
    \centering
    \includegraphics[width=\linewidth]{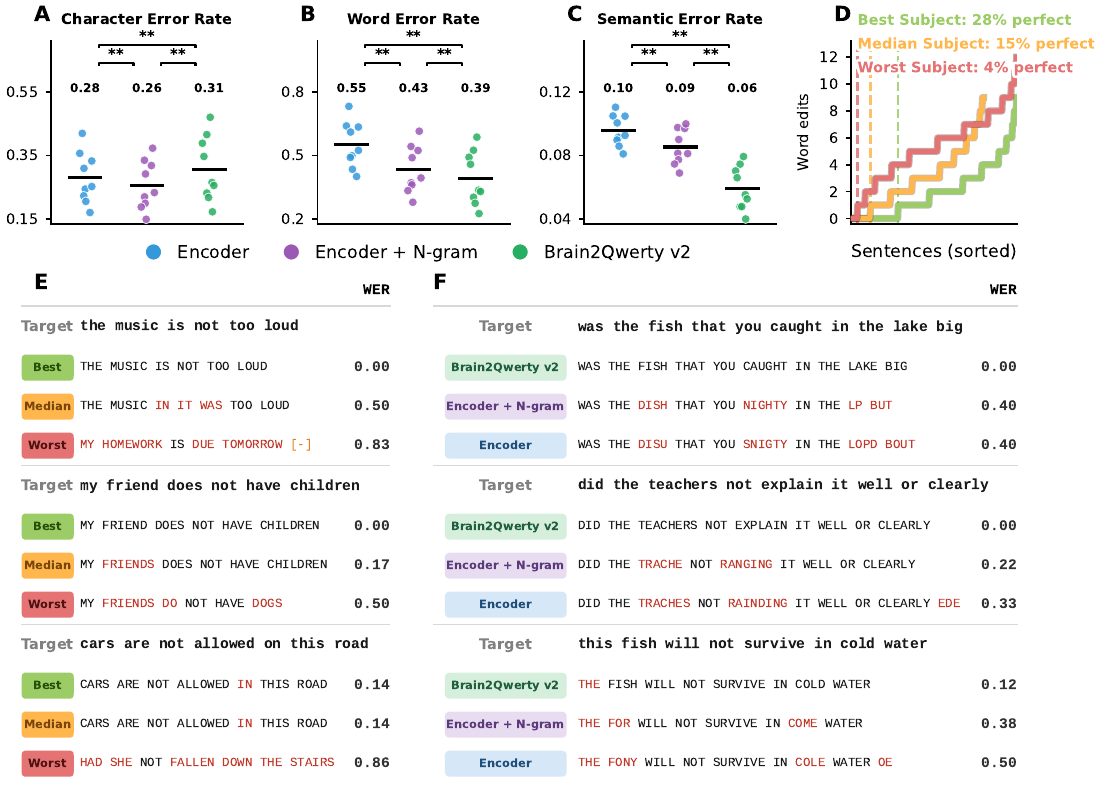} 
    \caption{\textbf{Brain2Qwerty v2 enables word- and meaning-level decoding from MEG.} \\ 
    \textbf{A--C.} Per-subject Character (\textbf{A}), Word (\textbf{B}), and Semantic (\textbf{C}) Error Rate for three decoders: Encoder -- MEG encoder greedy CTC predictions; Encoder + N-gram -- encoder predictions decoded with a 6-gram character language model; Brain2Qwerty v2 -- full pipeline conditioned on both CTC text and MEG-derived word embeddings. Each dot is one subject; the black bar is the cross-subject mean, printed above each column.
    \textbf{D.} Per-sentence word-edit count for the Best, Median, and Worst Subject, sorted ascending; colour-matched dashed lines mark each subject's perfect-decoded boundary ($\mathrm{WER} = 0$, annotated as ``\% perfect'').
    \textbf{E.} Decoded sentences from Brain2Qwerty v2 for the Best, Median, and Worst Subjects on three example targets.
    \textbf{F.} Decoded sentences from the three decoders (Brain2Qwerty v2, Encoder + N-gram, Encoder) on three example targets from the median subject (S01). In \textbf{A}--\textbf{B}, ground-truth sentences are shown next to ``Target''; word substitutions and insertions are highlighted in red and deletions are denoted by \texttt{[-]}.
    Statistical tests in \textbf{B}--\textbf{D}: two-sided paired Wilcoxon signed-rank across $n = 9$ subjects. Brackets show three adjacent decoder comparisons (lower level) plus the outer Encoder vs.\ Brain2Qwerty v2 comparison (upper level). Significance levels: ${*}\,p < 0.05$, ${**}\,p < 0.01$, ${***}\,p < 0.001$. All annotated comparisons reach $p = 0.0039$ ($**$), the minimum achievable for $n = 9$ paired observations.
    }
    \label{fig:fig3}
\end{figure}

\paragraph{Brain2Qwerty v2 outperforms baselines in word-level and semantic precision.}
We compare Brain2Qwerty v2 against two baseline approaches: (1) the Encoder alone without any language model correction, where the most likely keystroke class at each time step is selected from the CTC output logits, and (2) the Encoder + N-gram approach from Brain2Qwerty v1, in which the Encoder output is corrected by a N-gram language model~\citep{levy2025brain}. We evaluate all approaches on three complementary metrics: 1) Character Error Rate (CER), measuring character-level decoding accuracy, 2) Word Error Rate (WER), measuring word-level precision, and 3) Semantic Error Rate (SemER), quantifying semantic proximity between decoded and target sentences (Figure \ref{fig:fig3}A-C).

Brain2Qwerty v2 substantially outperforms both baselines on the WER and SemER metrics that are most relevant to successful communication. On WER, Brain2Qwerty v2 achieves a score of $0.39 \pm 0.04$, significantly lower compared to $0.55\pm 0.04$ for the Encoder alone ($p<0.005$) and $0.43 \pm 0.04$ for the Encoder + N-gram baseline ($p<0.005$) (Figure \ref{fig:fig3}B). On SemER, Brain2Qwerty v2 scores $0.059  \pm 0.005$, outperforming both the Encoder alone ($0.096 \pm 0.003, p<0.005$) and the Encoder + N-gram baseline ($0.085 \pm 0.004, p<0.005$) (Figure \ref{fig:fig3}C). These results reflect the LLM's capacity to recover globally coherent sentence structure from noisy neural inputs, as illustrated clearly in the cross-model example comparisons from the median subject (Figure \ref{fig:fig3}F). For sentences where the Encoder and N-gram outputs are fragmented and lexically incoherent, Brain2Qwerty v2 recovers the correct sentence or a semantically close approximation (Figure \ref{fig:fig3}F). Compared to the 0.92-0.94 WER on decoded perceived speech from fMRI by \cite{tang2023semantic}, our Brain2Qwerty v2 performance marks a significant improvement in decoding exact words from non-invasive neural recordings. More examples are displayed in Figure~\ref{fig:appendix-decoded-examples}.

Brain2Qwerty v2 does, however, incur higher CER ($0.31 \pm 0.03$) than both the Encoder alone ( $0.28 \pm 0.03, p<0.005$) and the N-gram baseline ($0.26 \pm 0.03, p<0.005$) (Figure \ref{fig:fig3}A). This is a consequence of autoregressive decoding using LLMs, which are trained to produce fluent sentences even when the encoder output and the MEG token signal quality are not sufficient for successful decoding, resulting in incorrect sentences that diverge substantially from the target at the character level. This failure mode of Brain2Qwerty v2 is qualitatively distinct from the local errors produced by the N-gram model. While Brain2Qwerty v2 produces either perfect or near-perfect decoding for the best subject, the worst subject's output can be a coherent but entirely different sentence (e.g., \emph{had she not fallen down the stairs} for the target \emph{cars are not allowed on this road}), incurring a large CER penalty with no corresponding gain in word or semantic accuracy. In contrast, the N-gram model consistently corrects local character sequences, leading to a lower CER, but fails to produce lexically correct sentences (Figure \ref{fig:fig3}F). Ultimately, because successful communication relies on meaning rather than strict character matching, Brain2Qwerty v2's robust improvements on WER and SemER establish its capacity to decode intelligible, meaning-driven sentences from non-invasive recordings.

\paragraph{Brain2Qwerty~v2 adapts a LLM to read MEG tokens.}
To establish that the LLM in Brain2Qwerty~v2 is reading the neural signal rather than refining the encoder's CTC predictions only with its language priors, we train an ablated version of Brain2Qwerty~v2 in which the LLM receives only the Encoder CTC output and no MEG tokens, and compare its performance against the full model and both baselines across all three metrics (Fig.~\ref{fig:neurotokens}). Removing the MEG tokens degrades every metric (CER $0.38\pm0.03$, WER $0.49\pm0.04$, SemER $0.067\pm0.004$; paired Wilcoxon $p \approx 0.004$ on all three), showing that the LLM is actively integrating the MEG-derived word embeddings into its predictions rather than treating them as a redundant copy of the CTC sequence. The two input streams thus play complementary roles---the CTC text anchors the LLM to a plausible linguistic prior, while the MEG tokens carry the residual neural information needed to push the decoding past that prior---and the full Brain2Qwerty~v2 performance is reached only when both are provided. These results establish that Brain2Qwerty~v2 is effectively a fine-tuned LLM that has learned to rely on brain activity to improve its predictions, and that this adaptation is obtained from as few as $\sim 2{,}700$ unique training sentences ($\sim 90$\,h of MEG), a corpus that is orders of magnitude smaller than what is typically used to fine-tune a language model.

\paragraph{Brain2Qwerty~v2 produces word-perfect sentences for a meaningful share of trials.} 
For our best subject, $28\%$ of test sentences are perfectly decoded (zero word edits) and $47\%$ are decoded within at most one word edit (Fig.~\ref{fig:fig3}D). The fraction of perfect decodings drops to $15\%$ for the median subject and $4\%$ for the worst subject, but the typical error mode remains a single substituted or missing word rather than a collapse to an unrelated sentence. Representative decodings (Fig.~\ref{fig:fig3}E) confirm this: even when individual words are mis-predicted, the meaning of the sentence is largely preserved on the best and median subjects, and the output remains a fluent, well-formed English sentence.

\subsection{From MEG to LLM: align by word, adapt by subject}

\begin{figure}[t!]
    \centering
    \includegraphics[width=\linewidth]{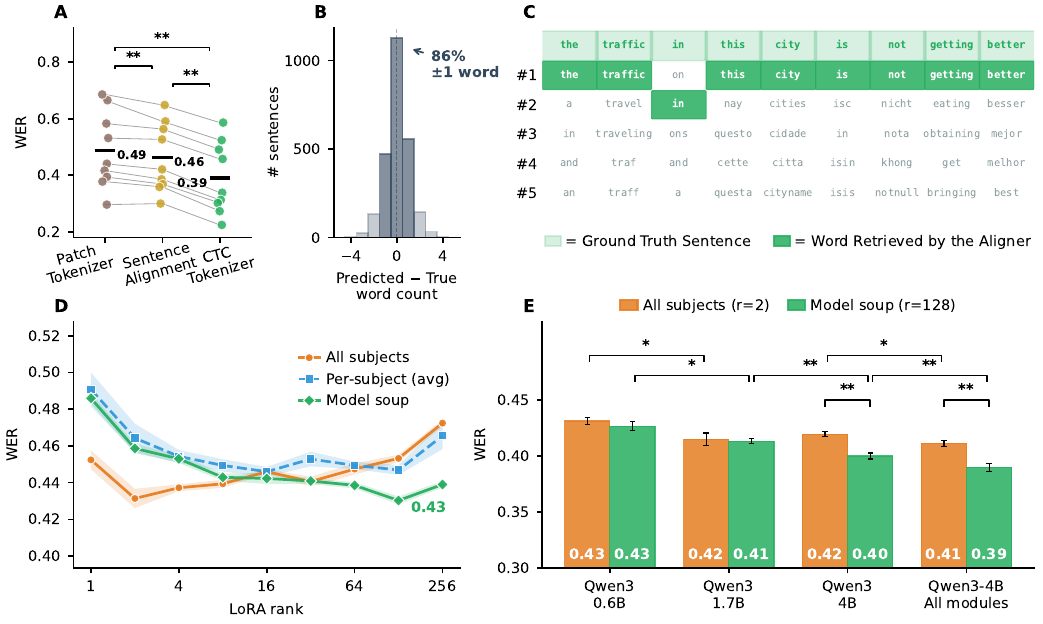}
    \caption{\textbf{CTC Tokenizer and tuned LLM adaptation cut WER by 20\% over baseline alignments.} \\
    \textbf{A.} Per-subject word error rate for three contrastive token alignment strategies: Patch Tokenizer (fixed number of patches), Sentence Alignment (one embedding per sentence), CTC Tokenizer (CTC-segmented embeddings based on the predicted space token).
    \textbf{B.} Word-count calibration of the CTC Tokenizer.
    \textbf{C.} Example of word-level token retrieval using the CTC Tokenizer strategy. Each column lists the top-5 tokens corresponding to different words of the Qwen3-0.6B vocabulary ranked by cosine similarity to the brain-derived word embedding. A green cell marks the rank at which the ground-truth word is retrieved.
    \textbf{D.} LoRA rank sweep for Qwen3-0.6B (attention-only targets, $\alpha = 2r$). Three training strategies are compared across LoRA ranks: All Subjects (single adapter for all subjects), Per-subject (independent adapters per subject), and Model Soup (uniform average of the per-subject model weights). Shaded bands show SEM across the 9 subjects, computed by subtracting each subject's mean across conditions before taking the standard deviation so that they reflect within-subject variability only.
    \textbf{E.} LLM backbone scaling and LoRA target-module ablation. Each group shows two bars: All Subjects at $r = 2$ (orange) and Model Soup at $r = 128$ (green). The first three groups use attention-only LoRA with increasing backbone size (Qwen3-0.6B, 1.7B, 4B); the fourth uses Qwen3-4B with LoRA extended to all seven linear projections. Error bars use the same SEM definition than in D.
    All metrics are sentence-level WER, averaged per subject and then across the 9 subjects. Brackets report paired two-sided Wilcoxon signed-rank tests ($^{*}\,p<0.05$, $^{**}\,p<0.01$, $^{***}\,p<0.001$); non-significant comparisons are unlabelled.
    }
    \label{fig:fig4}
\end{figure}


\paragraph{CTC tokenizer enables better alignment to the LLM semantic space.}
Our CTC tokenizer utilizes space keys to segment continuous MEG embeddings into word-like tokens. 
Because spaces are frequent (19\% of characters) and robustly predicted, this segmentation is accurate --- the predicted word count of 86\% of sentences fall within $\pm 1$ word of the ground truth (Fig.~\ref{fig:fig4}B). 
We compare this to two alternatives: sentence-level alignment (one embedding per sentence, as in~\citep{zhang2026crossspeciesneuralfoundationmodel}) and patch tokenization (a fixed number of equally-spaced chunks per sentence, inspired by vision transformers~\citep{dosovitskiy2021image}). Our CTC tokenizer reaches WER = 0.39$\pm$0.04, significantly better than sentence alignment (0.46, $\pm$ 0.04) and patch tokenization (0.49, $\pm$ 0.05) ($p < 0.005$; Fig.~\ref{fig:fig4}A). Single-word retrieval confirms the mapping: 8 of 9 words in a typical sentence are recovered at rank~1 (Fig.~\ref{fig:fig4}C). The CTC tokenizer outperforms alternatives, providing a generic technique for any CTC-based pipeline.

\paragraph{Treating each subject as a separate task lets LoRA scale without overfitting.}
LoRA adapters fine-tune a large pretrained LLM by training only a small
low-rank correction at each layer, controlled by one hyper-parameter
--- the rank $r$. Larger $r$ adds more trainable parameters and more
expressivity, but also more risk of overfitting --- a real concern
with our small dataset of $\sim$2.7K unique sentences.
We test three training strategies across $r \in \{1, \dots, 256\}$
(Fig.~\ref{fig:fig4}D). \emph{Joint LoRA} pools all nine subjects and
trains a single adapter on the combined data. \emph{Per-subject LoRA}
trains nine separate adapters, one per subject. \emph{Model
Soup}~\citep{wortsman2022modelsoups} also trains per-subject adapters
but then averages their weights uniformly into a single adapter that
is applied to every subject. The intuition is that each subject's MEG
signature is a different ``task''~\citep{wei2025optmerge}:
per-subject adapters specialize on matched neural data, and uniform
weight averaging yields a single model that generalizes across
subjects without ever training one heavy joint model.
The three strategies trace a clear pattern. Joint LoRA wins at
small ranks ($r = 2$, WER = 0.43); above that, the single
adapter has enough capacity to memorize the limited sentence pool and WER worsens. Per-subject LoRA and Model Soup see $\sim$10$\times$ less data each but never reach this overfitting regime: WER decreases smoothly with $r$, and Model Soup wins at
$r = 128$ (WER = 0.43). Model Soup also trains faster than the joint baseline, even at higher rank, as fitting nine small adapters is more efficient than training a single large pooled adapter. We ran additional comparisons between the two best configurations with larger LLMs and broader adapter targets (Fig.~\ref{fig:fig4}E). Model Soup keeps improving at every step --- 0.6B~$\rightarrow$~1.7B ($p = 0.020$), 1.7B~$\rightarrow$~4B ($p < 0.01$), attention-only~$\rightarrow$~all modules (where LoRA is applied) ($p < 0.01$)
--- while joint training plateaus near WER = 0.41. Model Soup is therefore the strategy that scales the best with LLM size. It gives a reusable recipe for plugging an LLM into a brain-to-text pipeline: each subject is treated as a distinct task, a separate low-rank adapter is trained per subject, and the resulting adapters are averaged in weight space.

\subsection{Auto Research}

\begin{figure}
    \centering\includegraphics[width=\linewidth]{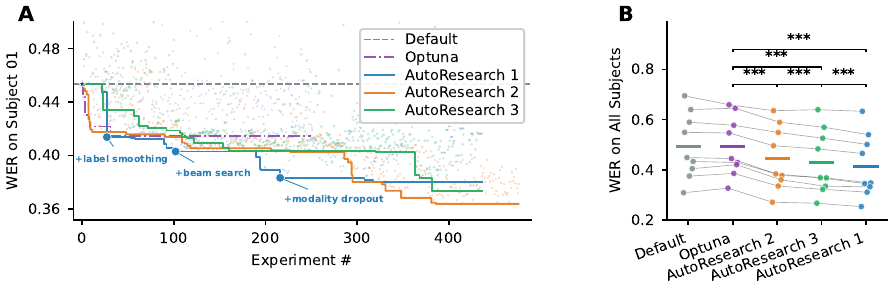}
    \caption{\textbf{Autonomous AI coding agents discover configurations that outperform classical optimization.} \\ 
    \textbf{A.}~Running-best validation WER on Subject~S01 over successive experiments. Each dot represents one training run; colored step functions track the cumulative best WER for each method. The dashed grey line indicates the default configuration with only 4 parameters: learning rate, batch size, weight decay and rank with reasonable values. Optuna search is denoted with purple (dash-dotted). It searches within the 4-parameters space. Three independent AutoResearch agents are in blue, orange and green. 
    \textbf{B.}~Test WER evaluated on all 9 subjects for each method's final configuration. Each dot represents one subject's mean WER; colored horizontal lines indicate the cross-subject mean. Methods are ordered by decreasing (worse) WER from left to right. Significance brackets show paired Wilcoxon signed-rank tests between Optuna and each AutoResearch agent ($^{***}$\,$p < 0.001$).}
    \label{fig:autoresearch}
\end{figure}

\paragraph{Approach.} 
We next investigate whether AI could not only be used for feature extraction (Encoder) and language modeling (LLM) but also for code development. For this, we employ an `Auto Research' inspired from Karpathy\footnote{\url{https://github.com/karpathy/autoresearch}}. 
Specifically, we task three independent agents (Cursor\footnote{\url{https://cursor.com/}} from Claude Opus 4.6~\citep{anthropic2026claudeopus46}) to minimize the validation Word Error Rate (WER) on a single subject, starting from a deliberately stripped-down configuration, where only four hyperparameters were explicitly specified: learning rate, weight decay, LoRA rank, and batch size, and let the agent iteratively change the codebase and grid search to explore new ideas. 
For comparison, we ran Optuna~\citep{akiba2019optuna} using Tree-structured Parzen Estimation (TPE) over the same four exposed parameters with an identical compute budget. The complete protocol is detailed in the Methods section.

\paragraph{Auto Research successfully improves our decoding pipeline.} 
The default configuration yielded a validation WER of 0.45 on Subject S01. As shown in Fig.~\ref{fig:autoresearch}A, Optuna quickly plateaued at a WER of 0.41 (8.6\% relative improvement), with subsequent TPE-guided trials unable to push the 4-parameter optimum further. In contrast, all three Auto Research agents continuously improved throughout their 10-round loops, achieving final validation WERs of 0.38, 0.36, and 0.37 (relative improvements of 16.1\%, 19.8\%, and 17.7\%, respectively). These gains materialized because the agents autonomously coded and tested parameters that Optuna, by construction, cannot access. 
To assess whether these improvements overfitted to Subject S01, we trained each best configuration on all subjects (Fig.~\ref{fig:autoresearch}B). Optuna's single-subject gain vanished during cross-subject evaluation (test WER = 0.493, $p = 0.88$ vs.\ default configuration). Conversely, all three Auto Research configurations maintained significant improvements over both the baseline and Optuna (all $p < 10^{-6}$), achieving test WERs of 0.42 (-16.0\%), 0.45 (-10.0\%), and 0.43 (-12.7\%). This demonstrates that the innovations discovered by the agents are genuinely beneficial across the broader population, rather than artifacts of single-subject tuning. All three agents independently converged on overlapping core strategies with Agent 1 producing the best overall configuration. Across the three agents, the most impactful changes implemented are:

\begin{itemize}
    \item Label smoothing: Discovered by all three agents in rounds 1--2, providing the largest single-round WER reduction (-0.04 for Agent 1).
    \item Modality dropout: Dropping CTC tokens during training forced the LLM to rely more on neural word embeddings rather than noisy CTC predictions, consistently yielding large gains in rounds 5--9.
    \item Beam search decoding of LLM output: Enabled multi-hypothesis decoding at test time, discovered in rounds 2--3.
    \item Contrastive alignment loss on the sentence level: Activated an auxiliary CLIP-style loss between neural and text sentence-level embeddings during LLM fine-tuning to complement the word-level alignment, discovered in rounds 7--8.
    \item Minimal prompts: Reduced the instruction prompt to its strict minimum (\texttt{CTC:}, \texttt{MEG:}, \texttt{Output:}) rather than using verbose task descriptions.
\end{itemize}

These techniques were all considered during the development of Brain2Qwerty v2, with the final configuration selected by the research team.  

\paragraph{Auto Research failed on open-ended optimization.} 
These results are, however, obtained within a deliberately constrained search space: the model architecture, data pipeline, loss formulation, and allowable runtime are fixed by the researcher, and only a limited set of training-time hyperparameters is exposed to the agents. When the same agents are instead given an open-ended objective: starting from the Brain2Qwerty v1 codebase and tasked with matching the performance of our final v2 pipeline on the EnglishBCBL dataset, they consistently fail: large, entangled code modifications caused the majority of subsequent SLURM jobs to crash before producing valid metrics, and on the rare occasions when a single launch does succeed, the agents tend to idle rather than iterate. 


\section{Discussion}
\paragraph{Summary.}
Brain2Qwerty v2 can accurately decode sentences solely from continuous non-invasive brain recordings by jointly training three deep learning modules, focusing on decoding character, word, and sentence-level representations, respectively.
This model achieves WER of 39\% on average across subjects, and 22\% WER for our best subject. This is a twofold improvement over the previous state-of-the art (52\% for the best subject in Brain2Qwerty v1~\citep{levy2025brain}) significantly narrowing the gap with invasive brain-to-text decoders \citep{willett2021handwriting,metzger2023neuroprosthesis}.

\paragraph{Semantic decoding.}
To our knowledge, Brain2Qwerty v2 is the first architecture that jointly trains the decoding of characters, words, and sentences. Most interestingly, semantic representations seem to be effectively leveraged by our model: the prompt ablation analyses indeed show that removing the `MEG tokens' from the LLM input degrades WER by 16\%. This finding is consistent with growing evidence that brain signals can be used to recover semantic information \citep{mitchell2008predicting,huth2016natural,broderick2022more,tang2023semantic,defossez2023decoding,dascoli2025towards,landau20252025}. In particular,~\citet{defossez2023decoding} demonstrated with contrastive alignment that perceived speech representations can be decoded from MEG signals, 
a result further extended to word embedding during reading and listening \citep{dascoli2025towards} and to language production tasks performed with intracranial and scalp EEG ~\citep{goldstein2023deep,liAssemblingMindsMosaic2026}. 
However, two elements should be stressed. 
First, Brain2Qwerty v2 per sentence decoding performance correlates significantly with the character-level representations (Figure \ref{fig:encoder_importance}A, B), indicating that the model remains grounded in character-level signal even when semantic representations are available. 
Second, there appears to be a trade-off between semantic and character-level decoding: 
Brain2Qwerty v2 does outperform an alternative model correcting decoded characters with an N-gram model on word-level and semantic metrics but it worsens CER. Similarly, 
when the encoder output is heavily corrupted, the LLM generates fluent but incorrect sentences (Figure \ref{fig:appendix-decoded-examples}). 
Depending on the condition (typing a password \emph{vs} responding to a dialogue), the decoding objective may need to be adjusted to best serve the patient’s needs.

\paragraph{Auto Research.}
The integration of autonomous AI agents into the Brain2Qwerty v2 development cycle marks a shift in the optimization of neural decoding pipelines. Motivated by the limitations of traditional Bayesian methods \citep{akiba2019optuna} -- which are restricted to predefined sets of parameters -- we here tasked AI agents to iteratively change our code base to invent novel, better architectures. This led to a substantial improvement of WER, and revealed the utility of sophisticated strategies, such as modality dropout and short prompts, which turned out to robustly generalize across all subjects (Figure \ref{fig:autoresearch}). Nevertheless, `Auto Research' is not (yet) replacing human researchers: Indeed, when tasked to autonomously find a good asynchronous model to decode brain activity, given Brain2Qwerty v1's baseline, the AI agents completely failed. These results suggest that while AI agents may serve as a powerful force multiplier, human research remains, for now, a critical part of the scientific process.

\paragraph{Inter-individual variability.}
Several limitations must be addressed before adapting this to the clinics. 
First, large inter-subject variability persists (N-gram CER: 17.1\%–41.0\%). Given that the final model performance is highly correlated with upstream encoder quality (Figure \ref{fig:encoder_importance}A, B), improving the encoder through cross-subject transfer or self-supervised pretraining is therefore a priority. 
Second, our study is conducted with healthy volunteers, effectively typing on a keyboard. It thus remains critical to demonstrate that this approach can be reliably adapted to patients, for whom the actual key presses will be missing, not just during inference, but also during training and/or finetuning.

\paragraph{Real-time latency.}
Currently, Brain2Qwerty v2 is not based on a causal architecture, and thus works with an entire sentence. This design choice implies that the real-time latency of this model is necessarily slow, and would not allow users to see the word they type before the end of the sentence. However, results in EMG have shown promising results in using a causal Conformer with low-latency to decode handwriting \citep{sivakumar2024emgqwerty}. A future step could be to investigate how to implement our model in a fully real-time low latency setting. 

\paragraph{Scaling laws.}
While Brain2Qwerty v2 narrows the gap with intracortical systems, a substantial performance difference remains. State-of-the-art invasive BCIs achieve below 2\% WER for typing~\citep{judeRestoringRapidNatural2026} and below 6\% CER for handwriting~\citep{willett2021handwriting}. Still, our scaling curves show no sign of plateauing at 90 hours of aggregated data (Fig \ref{fig:fig1}F), suggesting that data collection may be a simple lever for further improvement. Future validation of this scaling law is critical; if extended training on non-invasive MEG data can eventually obviate the need for neurosurgery, it would represent a transformative shift in patient care.

\paragraph{A path towards non-invasive BCI.}
This approach, based on a 306-sensor cryogenic MEG sensors, remains currently challenging to adapt to a clinical setup. We remain optimistic, however: 
first our analyses show a remarkably robust decoding with 25\% and 50\% of the MEG sensors (Table \ref{tab:sensor-ablation}). It will thus be important to later investigate whether an optimized selection of sensors could maintain a high performance, while minimizing  hardware constraints. 
Second, the MEG community continues to improve the development of optically-pumped MEG sensors~\citep{boto2018moving,schofield2022quantum}. These wearable devices work at room temperature and thus provide an important research avenue to transform the present laboratory proof-of-concept into a safe communication device for the clinics.
Ultimately, we hope these results will prompt a broader reconsideration of non-invasive BCI to offer a safe and scalable alternative to neural prostheses for restoring communication in individuals with speech or motor impairments.

\section{Methods}

\subsection{Experimental Protocol}

\paragraph{Participants}
A group of 9 healthy adult volunteers participated in the study at the Basque Center on Cognition, Brain and Language (BCBL) in Spain. The cohort consisted of 50\% men and 50\% women, with a mean age of 34.6 years (range 23-56). All participants were right-handed and proficient typists able to type without looking at their hands. Only individuals whose typing accuracy met or exceeded 80\% were included. All participants were native English speakers with no self-reported history of neurological or psychiatric conditions. Brain activity was captured using MEG, with 10 recording sessions, each lasting for approximately 1 hour. All participants provided written informed consent and received 12 euros per hour as compensation (plus extra 200 euros upon completion). The study received approval from the local ethics committee. Prior to analysis, all records were de-identified and anonymized such that no personally identifiable information was accessible. 

\paragraph{Neuroimaging Devices}
This dataset has been collected with a Megin system with 306 channels (102 magnetometers and 204 planar gradiometers). We record at a sampling rate of 1kHz, with an online high-pass filter set at 0.1Hz and a low-pass filter at 330Hz. As in \cite{levy2025brain}, we used a custom MR-compatible QWERTY keyboard from HybridMojo (LLC) with modified non-ferromagnetic springs.

At the end of one of the sessions, each subject underwent an anatomical MRI scan for 15 minutes, if they were not already available in the database of the Basque Center on Cognition Brain and Language (BCBL). For this, we used a 3-T SIEMENS Prisma-fit scanner (Siemens Medical Solutions), with a 64-channel head coil. High-resolution T1-weighted anatomical images were obtained with the following acquisition parameters: TR = 2530 ms, TE = 2.36 ms, flip angle = 7°, Field of view = 256 mm, voxel resolution = 1 mm3; 176 slices.

\paragraph{Task}
Participants were seated facing a projected screen (100 cm from their eyes), with the keyboard positioned on a stable platform. The MEG sensors were kept at a fixed distance of 70 cm from the keyboard, allowing participants to type in a natural and comfortable posture. Each trial followed a three-phase structure: listening, wait for cue, type. First, a sentence audio was played through MEG-compatible headphones worn by the participants. Second, a fixation cross was shown on screen for 1.5 seconds. Finally, the offset of the fixation cross marked the beginning of the typing phase, during which no letters were displayed on screen. Participants were instructed to type the sentences they heard as accurately as possible without using backspace while fixated on the rotating black square in the center of the screen. To avoid eye movements being driven by linguistic content — as is typically the case in standard left-to-right reading — minimal visual feedback was provided in the form of a small square at the center of the screen that rotated clockwise by 10 degrees with each keystroke. Each recording session comprised 16 blocks of 16 sentences each. The first 4 sentences of every session served as practice trials and were distinct from the 2560 unique sentences used in the protocol. During the first two practice sentences, participants received full visual feedback while typing, while the remaining two were used to familiarize them with the minimal feedback condition.

\paragraph{Sentences}
Each session consisted of 256 sentences, and each participant participated in 10 sessions in total. The stimuli were drawn from an initial pool of 20,000 simple English sentences generated using the Llama 4 model \citep{meta2025llama4}. Prior to selection, this pool was filtered to remove special characters and exclude any contractions (e.g., \textit{don't} for \textit{do not}). From this refined pool, 2,560 unique sentences were randomly selected to be shared across participants. Ultimately, the final analysis encompasses 2,724 unique sentences due to the inclusion of pilot sessions conducted with the first participant.

\paragraph{MEG Preprocessing}
MEG signals were acquired from whole-head sensor arrays and preprocessed offline. Raw data were bandpass filtered between 0.5 and 45 Hz and a notch filter at 50 Hz was applied to suppress power-line interference. Signals were downsampled to 100 Hz. Per-channel amplitudes were normalized using a RobustScaler (median and interquartile-range statistics estimated per recording), and values exceeding ±5 robust standard deviations were clamped. No signal-space projection was applied.

\paragraph{MEG Source reconstruction}
To estimate the cortical sources of the MEG signal, we performed source reconstruction using the dynamic statistical parametric mapping (dSPM) method \citep{dale2000}, as implemented in MNE-Python \citep{gramfort2013}. 
Individual anatomical MRI scans were processed with FreeSurfer's `recon-all` pipeline. Coregistration between MEG and MRI data was performed using digitized head shape and fiducial points (nasion and preauricular points), enabling precise alignment of MEG sensor locations with individual cortical anatomy.
A single-layer boundary element model (BEM) was constructed from each subject's inner skull surface using the watershed algorithm, appropriate for MEG-only data. The cortical source space was defined on the white matter surface at 'oct6' spacing (\~4098 dipoles per hemisphere). Forward models were computed per participant using the individual BEM, source space, and coregistration transform.
The noise covariance matrix was estimated from the baseline of sentence-onset epochs (from -500 to -50\,ms) across all recording sessions per participant, using the empirical (sample) covariance estimator. Per-session covariance matrices were then combined using a naive-weighted average. Epochs were bandpass-filtered between 0.1 and 40~Hz using a zero-phase IIR (Butterworth) filter prior to covariance estimation. The inverse operator was computed with standard parameters, namely: loose orientation constraint ($\text{loose}=0.2$) and depth weighting ($\text{depth}=0.8$). The dSPM solution was applied with a signal-to-noise ratio of $\text{SNR}=3$ ($\lambda^2 = 1/\text{SNR}^2$). Individual source estimates were morphed to the \textit{fsaverage} template brain and grand-averaged across participants per condition. For the final plotting, we exclude three participants who had poor digitization. 

\paragraph{Text processing}
Each sentence was encoded as a character-level integer sequence, which is converted from the target sentence. The output vocabulary comprises 28 classes: a CTC blank token (index 0), the 26 lower-case English letters (indices 1--26) and a space token (index 27). Each sentence is assigned a unique identifier formed by concatenating its trial and session ID, ensuring unambiguous cross-referencing across subjects and sessions.

\paragraph{Train/validation/test splits}
Data were partitioned at the level of unique sentence texts using a deterministic hash-based splitter with an 80/10/10 ratio for training, validation, and test sets respectively. Because the assignment is computed from a hash of the sentence text alone, the split is stable regardless of which subjects are included in the query. All events sharing the same sentence text (across subjects and sessions) are assigned to the same partition, guaranteeing zero text leakage between splits.

\subsection{Decoding pipeline}

\paragraph{Data augmentation}

MEG signals were segmented from 400 ms before sentence onset  (first key pressed) through sentence offset (last key released), with the segment duration extended by a uniformly sampled buffer of 400--500 ms to capture post-completion neural activity. During training, temporal jittering was applied by randomly cropping up to 400\,ms from the pre-onset baseline, varying the effective alignment between neural signals and sentence onset across epochs.

During training, each MEG segment underwent a stochastic augmentation pipeline. A per-channel constant offset sampled from $\mathcal{N}(0, 0.3)$ was added to simulate slow-drift artifacts. Time masking was applied along the temporal axis with a maximum mask length of 50 frames and application probability of 0.2, and an independent channel mask of maximum width 400 was applied along the sensor axis in a procedure analogous to SpecAugment~\citep{park2019specaugment}. Temporal stretch augmentation rescaled trial duration by a factor drawn uniformly from $[0.8, 1.2]$ via linear interpolation. 

\textbf{Architecture}

\emph{Encoder.} MEG data were processed by a hierarchical encoder consisting of two stages: (i) a BrainModule, consisting of a convolutional feature extractor with spatial channel merging inspired by~\citet{defossez2023decoding}, and (ii) a Conformer sequence model~\citep{gulati2020conformer}.

\begin{itemize}
    \item \textbf{Spatial channel merging.}~Because MEG sensors are distributed in three-dimensional space and the number of active channels varies across sessions, raw per-channel time-series were first projected into a fixed-dimensional spatial representation. Channel coordinates were encoded as two-dimensional Fourier features with a total embedding dimension of 2,048 and passed through a learnable module that mapped the variable-length sensor array onto 270 virtual channels per time step. A per-subject affine layer, conditioned on a subject index, was applied to capture individual sensor geometry without requiring explicit sensor-level co-registration.
    \item \textbf{Convolutional encoder.}~The spatially fused representation ($B \times 270 \times T$, where $B$ is the batch size and $T$ is the number of time frames at 100~Hz) was processed by a four-layer dilated convolutional network (hidden dimension 1,500; kernel size 5; dilation period 3). Each layer included GELU activations, batch normalization, residual skip connections (scaled by 0.1), and dropout (input dropout 0.2; convolutional dropout 0.5). An initial linear projection reduced the channel dimension to 512 before the convolutional stack.
    \item \textbf{Temporal downsampling.}~A strided 1D convolution (kernel size 16, stride 4) reduced the temporal resolution by a factor of four, yielding approximately one frame per 40 ms.
    \item \textbf{Conformer.}~The downsampled sequence was processed by a four-layer Conformer \citep{gulati2020conformer} with model dimension 1,024, four attention heads, feed-forward dimension 1,024, and a depthwise convolution kernel of width 17. Group normalization was applied within each convolutional sublayer, and dropout of 0.3 was used throughout. We used the \textit{torchaudio} \citep{yang2021torchaudio} implementation.
\end{itemize}

\emph{Character-level CTC decoding.}
The Conformer output was projected by a linear head to the 28-class character vocabulary. The model was trained with the Connectionist Temporal Classification (CTC) loss~\citep{graves2006connectionist}. An auxiliary CTC head was attached after the convolutional encoder (after temporal downsampling) and trained jointly. The composite loss was:

$$\mathcal{L} = (1 - \alpha)\,\mathcal{L} _{CTC}^\text{final} + \alpha\,\mathcal{L} _{CTC}^\text{aux}, \quad \alpha = 0.7$$

giving the auxiliary branch the dominant gradient signal during early training while the Conformer refines the final representation. The auxiliary loss is implemented as in \citet{nozakiRelaxingConditionalIndependence2021}.


\emph{Neuro-conditioned language model (NeuroLLM).}
The frozen CTC encoder was coupled with a causal language model (Qwen3-4B~\citep{qwen3technicalreport}), fine-tuned via Low-Rank Adaptation~\citep{huLoRALowRankAdaptation2021} (LoRA per-subject then model soup; rank~128, $\alpha = 256$, dropout~0.0) applied to all linear projection matrices.

\begin{itemize}
    \item \textbf{Word-level segmentation and alignment.}~Neural word representations were derived by segmenting the Conformer output at frame positions where the CTC greedy path emitted a space token. Each word segment (including blank frames) was passed through a two-layer MLP applied per frame and then mean-pooled along the time dimension into a single neural word embedding. A linear adapter projected the encoder dimension to the LLM hidden dimension when the two differed.

    Because CTC-based segmentation may produce a different number of neural
    word tokens than the number of words in the target sentence, a
    within-sentence alignment step is required before word-level contrastive learning.
    For each sentence, a cosine-distance cost matrix is computed between the
    $N$ neural word embeddings and the $M$ target word embeddings (obtained
    from the LLM's input embedding layer). Hard DTW~\citep{sakoe1978dynamic}
    recovers a monotonic alignment path through this matrix, from which
    one-to-one matched pairs are extracted (one target per neural token).
    Matched pairs from all sentences in the batch are then concatenated,
    $\ell_2$-normalised, and used to compute a SigLIP
    loss following
    \citet{zhai2023sigmoid,dascoli2025towards}: for every pair
    $(i,j)$ of matched embeddings across the batch, a sigmoid binary
    cross-entropy is applied to the scaled cosine similarity
    $\tau\,\langle \hat{\mathbf{w}}_i, \mathbf{w}_j \rangle + b$, where
    $\tau$ and $b$ are learnable scalars. The target label is~$1$ whenever
    the ground-truth embeddings $\mathbf{w}_i$ and $\mathbf{w}_j$ have
    cosine similarity $\geq 0.999$ (i.e.\ they represent the same word),
    and~$0$ otherwise. This duplicate-aware labelling prevents the same
    sentence appearing for multiple subjects from creating false negatives.
    
    
    \item \textbf{Prompt construction and LLM.}~The LLM input was constructed as: [\texttt{CTC:}~$\|$~CTC-decoded text tokens~$\|$~\texttt{\textbackslash nMEG:}~$\|$~MEG token embeddings~$\|$~\texttt{\textbackslash nOutput:}]. During training, modality dropout independently zeroed random token positions in the MEG token embeddings and CTC text tokens (both at rate\,0.1), encouraging robust conditioning on either modality alone. The LLM was trained with cross-entropy loss and label smoothing of 0.02 to decode the target sentence autoregressively, conditioned on both the CTC-decoded text and the neural word embeddings.
    
    \item \textbf{Model soup.}~Rather than training a single LLM on all subjects jointly, we trained one LoRA adapter per subject independently, then uniformly averaged their best-checkpoint state dictionaries into a single model-soup checkpoint~\citep{wortsman2022modelsoups}.
    
    \item \textbf{Training regimes.}~Two training regimes were explored:
    
    \emph{(i)~End-to-end staged training.}~The full CTC\,+\,Contrastive\,+\,LLM pipeline was trained jointly over 275~epochs in three phases: CTC only (epochs\,0--149), CTC\,+\,contrastive alignment (epochs\,150--224, weight $\alpha = 0.1$), and CTC\,+\,contrastive\,+\,LLM cross-entropy (epochs\,225--274, weight $\beta = 0.01$). The overall loss is
    \[
        \mathcal{L} = (1-\alpha-\beta)\,\mathcal{L}_{\text{CTC}} + \alpha\,\mathcal{L}_{\text{Contrastive}} + \beta\,\mathcal{L}_{\text{CE}}\,,
    \]
    where inactive terms are dropped and active weights renormalised to sum to~1.
    
    \emph{(ii)~Standalone LoRA fine-tuning.}~Alternatively, the CTC encoder and pre-trained word projector were frozen from the best contrastive checkpoint, and only the LoRA-adapted LLM was optimised for 30~epochs on a single GPU. This lightweight regime enabled rapid iteration over LoRA configurations, LLM sizes, and the model-soup strategy described above, and yielded our best overall results.
    
    \item \textbf{Generation.}~At inference, beam search was used with beam size\,16, a maximum of 60~new tokens, and length penalty\,0.2.
\end{itemize}

\paragraph{Training procedure and compute}
We build our data loading and training pipeline using the neuralset \citep{king2026neuralset} and neuraltrain \citep{dAscoli2026TribeV2} libraries.
All models were trained with AdamW~\citep{loshchilov2017decoupled} (learning rate $8 \times 10^{-4}$, weight decay $10^{-3}$) using a linear warm-up of 500 steps (start factor 0.01) followed by a Cosine Annealing schedule. Training was conducted in BF16 mixed precision with gradient clipping at a global norm of 1 and gradient accumulation over 2 micro-batches, yielding an effective batch size of $64 \times 2 \times 8 = 1024$ samples across 8 GPUs. Validation and test batch sizes were 128 and 8, respectively. For encoders used for the LoRA fine-tuning experiments, the best checkpoint was selected by validation CER, with early stopping patience of 50 epochs. For the end-to-end staged pipeline (CTC\,+\,Contrastive\,+\,LLM), training ran for 275 epochs on $8 \times$ A100 80\,GB GPUs with no early stopping. The full training procedure is running in 19.5 hours.

\paragraph{Encoder+N-gram}
At inference, the per-frame log-softmax output of the final CTC head was decoded by a lexicon-free character-level beam search integrated with a 6-gram character language model trained on the WikiText-103 corpus~\citep{merity2017pointer} (KenLM binary format)~\citep{KenLM}. Decoding hyperparameters were: beam size 50, language-model weight 2.0, insertion bonus 0.0, and a maximum of 5 non-blank labels per time step. The highest-scoring beam hypothesis was retained as the final prediction.

\textbf{Evaluation Metrics}

We use the following evaluation metrics: 
\begin{itemize}
    \item Character error rate (CER): Levenshtein edit distance between predicted and ground-truth character sequences, normalised by the ground-truth length. 
    \item Word error rate (WER): Levenshtein edit distance computed at the word level after whitespace tokenisation, normalised by the number of ground-truth words. 
    \item Semantic error rate (SemER): the $\ell_2$ distance between $\ell_2$-normalised, mean-pooled hidden states of a frozen RoBERTa-large model~\citep{liu2019roberta} applied to the predicted and reference sentences.
\end{itemize}

Both CER and WER were implemented via the \texttt{SequenceMatcher} from the \texttt{edit-distance} library. SemER was implemented using the \texttt{transformers} library from HuggingFace. We performed two-sided Mann-Whitney U test for significance testing to compare between experimental conditions.

\subsection{Auto Research}
We developed a protocol for autonomous hyperparameter discovery using AI coding agents. Three independent agents (Cursor \footnote{\url{https://cursor.com/}} powered by Claude Opus~4.6 \citep{anthropic2026claudeopus46}) ran concurrently. Each had full filesystem and terminal access to a dedicated git worktree on an isolated branch. They shared a common codebase and pre-trained checkpoints but maintained separate configuration files and results directories. Agents operated strictly without access to each other's branches. The parent branch configuration was intentionally minimal, exposing only four hyperparameters (learning rate, weight decay, LoRA rank, batch size), the model identity (Qwen3-0.6B with LoRA), and architectural constants (encoder checkpoint, projector type).

Each agent received the following prompt: \\ 

\begin{mdframed}[
  linewidth=0.5pt,
  linecolor=gray,
  backgroundcolor=gray!10, 
  innertopmargin=8pt,
  innerbottommargin=8pt,
  innerleftmargin=10pt,
  innerrightmargin=10pt,
  roundcorner=3pt,
  font=\small\ttfamily\raggedright 
]
\textrm{\textit{Session variables:}} SUBJECT=0, SEED=42, TAG=AUTO\_N\\[4pt]
\textrm{\textit{Context (re-read at every round):}}\\
You are optimising a brain-to-text decoding pipeline. The encoder is pre-trained and frozen. Your job is to find the best LLM fine-tuning configuration that minimises \\ WER on validation data. Prioritise NOVEL ideas over refining known parameters. \\ A great round introduces a genuinely new concept that boosts the performance. \\ Each new idea should be implemented as a sweepable parameter that can be toggled in \\ the config.

\textrm{\textit{Rules:}}\\
-- Poll SLURM every 5\,min; never idle\\
-- Each round: EXACTLY 50 jobs testing $\geq$\,4--5 independent ideas\\
-- 45-min SLURM timeout; cancel hung jobs immediately\\
-- Never delete files; never inspect test predictions\\
-- Use only subject 0; optimise on validation metrics only\\[4pt]
\textrm{\textit{Budget:}} 10 rounds $\times$ 50 jobs = 500 total runs\\[4pt]
\textrm{\textit{Per-round loop:}}\\
1. Re-state the goal\\
2. Design a factorial grid (exactly 50 jobs, $\geq$\,4 orthogonal ideas)\\
3. Launch $\to$ poll $\to$ scan results\\
4. Integrate winning values into the running-best configuration\\
5. Plot evolution; proceed to next round\\[4pt]
\end{mdframed}

\paragraph{Experiment tracking.}
A custom append-only tracker scanned validation predictions from completed runs, computed the mean WER, and updated a cumulative log. Improvements were recorded only when the WER strictly decreased. To enforce the test-set holdout at the code level, the tracker was hardcoded to access only validation data, and agents were explicitly instructed never to inspect test predictions. An evolution plot was generated after every round to guide the agent. Cross-subject evaluation occurred only once, after all 10 rounds, using each agent's final configuration trained on all 9 subjects.

\paragraph{Grid launcher.}
Because agents tend to underutilize allocated capacity, a custom SLURM launcher enforced exactly 50 jobs per round, raising an error if the combinatorial grid product deviated. The launcher managed infrastructure defaults (single V100 GPU, 45-minute timeout, disabled checkpoints, early stopping with \texttt{patience}=3 and \texttt{n-epochs}=20) and assigned experiment identifiers. Each agent executed 10 rounds (500 jobs total). Total wall-clock time per agent was approximately 6--8 hours, dictated primarily by cluster queue times.

\paragraph{Optuna baseline.}
For comparison, we executed an Optuna TPE search (v4.8, \texttt{multivariate=True}, \texttt{group=True}, \texttt{constant-liar=True}, \texttt{n-startup-trials}=32) over the baseline's four hyperparameters using an identical compute budget. To match the agents' parallelism, 500 trials were distributed across 10 batches of 50, updating the TPE sampler between batches.

\section{Acknowledgements}
The authors would like to thank Maite Kaltzakorta and Manex Lete.
This research is supported by the Basque Government through the BERC 2022-2025 program and Funded by the Spanish State Research Agency through BCBL Severo Ochoa excellence accreditation CEX2020-001010/AEI/10.13039/501100011033.
Parts of this research were carried within the European Union's Horizon 2020 research and innovation programme under the Marie Skłodowska-Curie grant agreement No 945304 - Cofund AI4theSciences hosted by PSL University.





\clearpage
\bibliography{main}

\newpage
\appendix
\renewcommand{\thefigure}{S\arabic{figure}}
\setcounter{figure}{0}

\section*{Appendix}

\subsection{MEG source reconstruction over time}

\begin{figure*}[h]
    \centering
    \includegraphics[width=\textwidth]{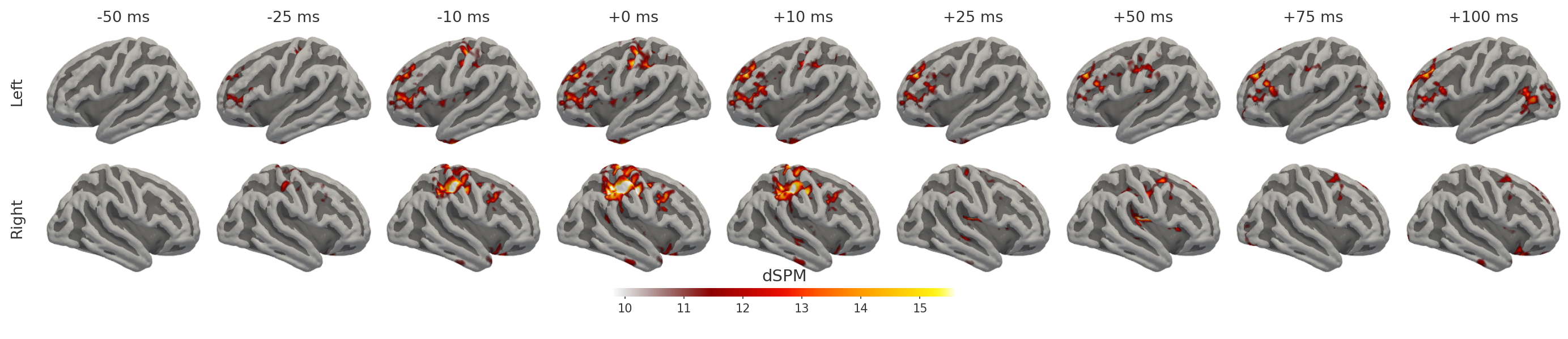}
    \caption{%
        \textbf{MEG sources over time relative to keystroke onset.} 
        }
    \label{fig:source_times}
\end{figure*}
\subsection{Clustering analysis of embeddings from BrainModule and Conformer}
\begin{figure*}[h!]
    \centering
    \includegraphics[width=\textwidth]{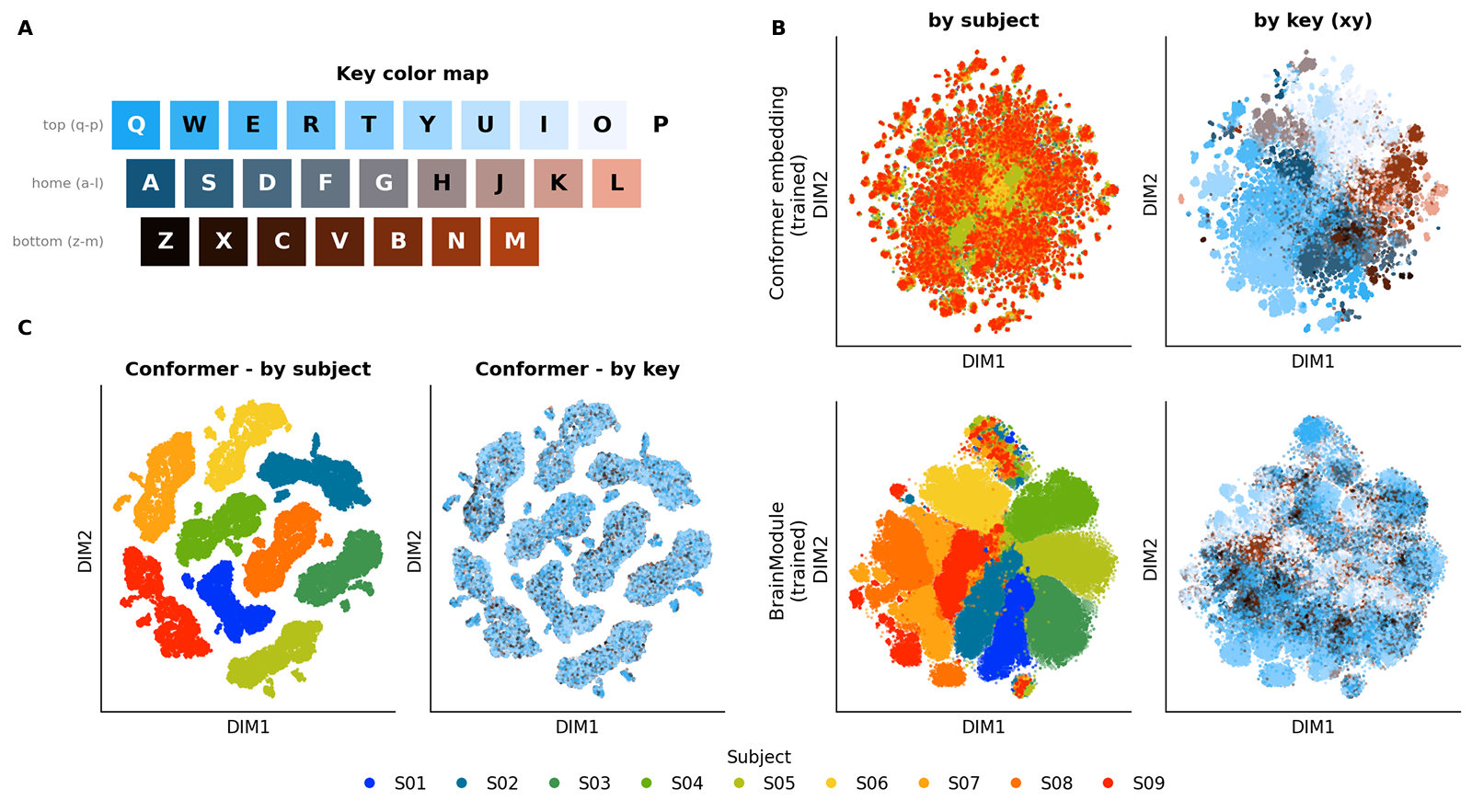}
    \caption{%
        \textbf{tSNE clustering.} \\ 
        \textbf{A. }Keyboard color map based on key location.
        \textbf{B. }tSNE clustering of key representations taken from the last layer of the Conformer of an untrained MEG Encoder, colored by subject (left) and key (right).
        \textbf{C.} tSNE clustering of key representations taken from the BrainModule (bottom row) and last layer of the Conformer (top row) of our trained MEG Encoder, colored by subject (left) and key (right).
        }
    \label{fig:tsne}
\end{figure*}

We perform a clustering analysis of the keystroke representations to interpret the embedding space learned by the Encoder. Representations are taken after the temporal down-sampling module in the BrainModule and the Conformer respectively in a trained Encoder. We also take the Conformer last layer embedding from an untrained model. We define the representation of a key as the embedding vectors of the time steps where a CTC emission labels a key class (not blanks nor space key). If multiple time steps continuously emit the same keystroke, we average across those time steps to create a single embedding. 

We apply tSNE clustering on the key representations taken from the test set sentences. Results show a clear difference in how the embedding space is organized between the BrainModule embeddings and the Conformer embeddings (Figure~\ref{fig:tsne}B). The key representations after the BrainModule cluster by participant, showing that participant-specific signatures remain in the key embeddings. However, the representations after the Conformer clearly cluster by key class, reflecting the physical layout of the keyboard. Crucially, this structure is learned from training, as seen in the embedding space of the untrained Conformer, where the key representations clearly cluster by subject (Figure~\ref{fig:tsne}C).

\subsection{Multi-subject training and cross-subject adaptation}

\begin{table}[h]
\centering
\caption{Effect of including the target subject in pretraining and of training on multiple subjects. \textbf{Per-subject} trains the full pipeline from scratch on the target subject only. \textbf{Joint training} trains from scratch on all subjects jointly. \textbf{LOO + finetune} pretrains on N-1 other subjects, then finetunes on the target with the Conformer frozen.}
\label{tab:xsub}
\begin{tabular}{lcccccc}
\toprule
 & \multicolumn{2}{c}{Per-subject} & \multicolumn{2}{c}{LOO + finetune} & \multicolumn{2}{c}{Joint training}\\
\cmidrule(lr){2-3} \cmidrule(lr){4-5} \cmidrule(lr){6-7}
Subject & CER & WER & CER & WER & CER & WER \\
\midrule
Best & 0.312 & 0.383 & 0.252 & 0.328 & 0.170 & 0.226 \\
Median & 0.530 & 0.665 & 0.475 & 0.586 & 0.368 & 0.478 \\
Worst & 0.707 & 0.906 & 0.566 & 0.683 & 0.482 & 0.614 \\
\bottomrule
\end{tabular}
\end{table}

To test how decoder performance depends on the training data composition,
we compare three regimes on the best, median, and worst subjects. In the per-subject regime, the full pipeline is trained from scratch on the target subject's data alone (4$\times$ the joint-training schedule to match the number of optimisation steps). In the joint regime, our chosen pipeline, we trained from scratch and evaluated per subject. In the leave-one-out + finetune regime, the pipeline is pretrained on the eight other subjects and then finetuned on the held-out subject with the Conformer frozen and the same training schedule. Performance follows the same ordering on all three subjects: per-subject worse than LOO + finetune worse than joint training. WER goes from 38.3\% (per-subject) to 32.8\% (LOO + finetune) to 22.6\% (joint) on best; from 66.5\% to 58.6\% to 47.8\% on median; from 90.6\% to 68.3\% to 61.4\% on worst (Table~\ref{tab:xsub}). Joint training on multiple subjects contributes meaningfully to final performance. Single-subject training is consistently the weakest, even with matched compute, but joint training is not strictly required: a single finetuning pass on a held-out subject closes most of the gap, supporting deployment to new subjects without retraining the base pipeline from scratch.

\subsection{Sensors ablation}
\label{appx:sensor-ablation}

\begin{table}[h]
    \centering
    \caption{\textbf{Sensor-count ablation.} Character (CER), Word (WER), and Semantic Error Rate (SemER) when the MEG input is randomly subsampled at training time. Values are $\mathrm{mean} \pm \mathrm{SEM}$ across 4 sensor-selection seeds; metrics are sentence-level, then averaged per subject, then averaged across subjects.}
    \label{tab:sensor-ablation}
    \begin{tabular}{cccc}
        \toprule
        Fraction & CER & WER & SemER \\
        \midrule
        25\% & $0.432 \pm 0.007$ & $0.547 \pm 0.009$ & $0.0735 \pm 0.0009$ \\
        50\% & $0.385 \pm 0.005$ & $0.490 \pm 0.006$ & $0.0684 \pm 0.0004$ \\
        75\% & $0.367 \pm 0.003$ & $0.467 \pm 0.003$ & $0.0662 \pm 0.0003$ \\
        \bottomrule
    \end{tabular}
\end{table}

Whole-head MEG arrays such as the 306-channel Megin system used
in the main experiments are bulky, cryogenically cooled, and expensive for clinical or consumer deployment. Optically pumped magnetometers (OPMs) which operate at room temperature in
flexible helmets are emerging as a practical alternative. They are typically composed of 50--150 sensors \citep{boto2018moving, schofield2022quantum}. Whether such low-channel arrays can support the kind of end-to-end sentence decoding presented in this paper is a very important question for the next generation of non-invasive brain-computer interfaces. We ran a controlled ablation to estimate the headroom.

For each keep-fraction in \(\{0.25, 0.50, 0.75\}\) we randomly
subsampled the 306 MEG channels down to \(\{76, 153, 230\}\) sensors
respectively. We retrained the entire Brain2Qwerty~v2 pipeline with
one LoRA for all subjects, rank=2 for simplicity. With the full
sensor array this configuration reaches a WER of \(0.433\)
(Figure~\ref{fig:fig4}); for each subsampled fraction we trained
\(n=4\) independent sensor-selection seeds, fixing the model seed so
that variance within a fraction reflects only the choice of sensors
retained.
Performance degrades smoothly (Table~\ref{tab:sensor-ablation}) and
monotonically with sensor count. Dropping the first 76 channels (from 306 to 230) costs only \(+3.4\) pp WER over the full-array baseline, dropping another 77 (from 230 to 153) costs \(+2.3\) pp, while dropping the final 77
(from 153 to 76) costs \(+5.7\) pp. SEMs are tight (always
$\leq 1$ pp on WER). The trend is not driven by which specific channels happened to survive subsampling. Sufficient information can be recovered even at low sensor counts. Crucially, an OPM-class helmet with on the order of \(150\) sensors loses only $\sim 5.7$ pp WER versus the full 306-channel baseline. Once paired with the same end-to-end decoding pipeline, such low-channel systems should recover most of the performance reported in the main paper, opening the door to non-invasive sentence decoding in deployable, room-temperature systems.

\clearpage
\subsection{Decoder behaviour across difficulty bands}
\begin{figure}
    \centering
    \includegraphics[width=\textwidth]{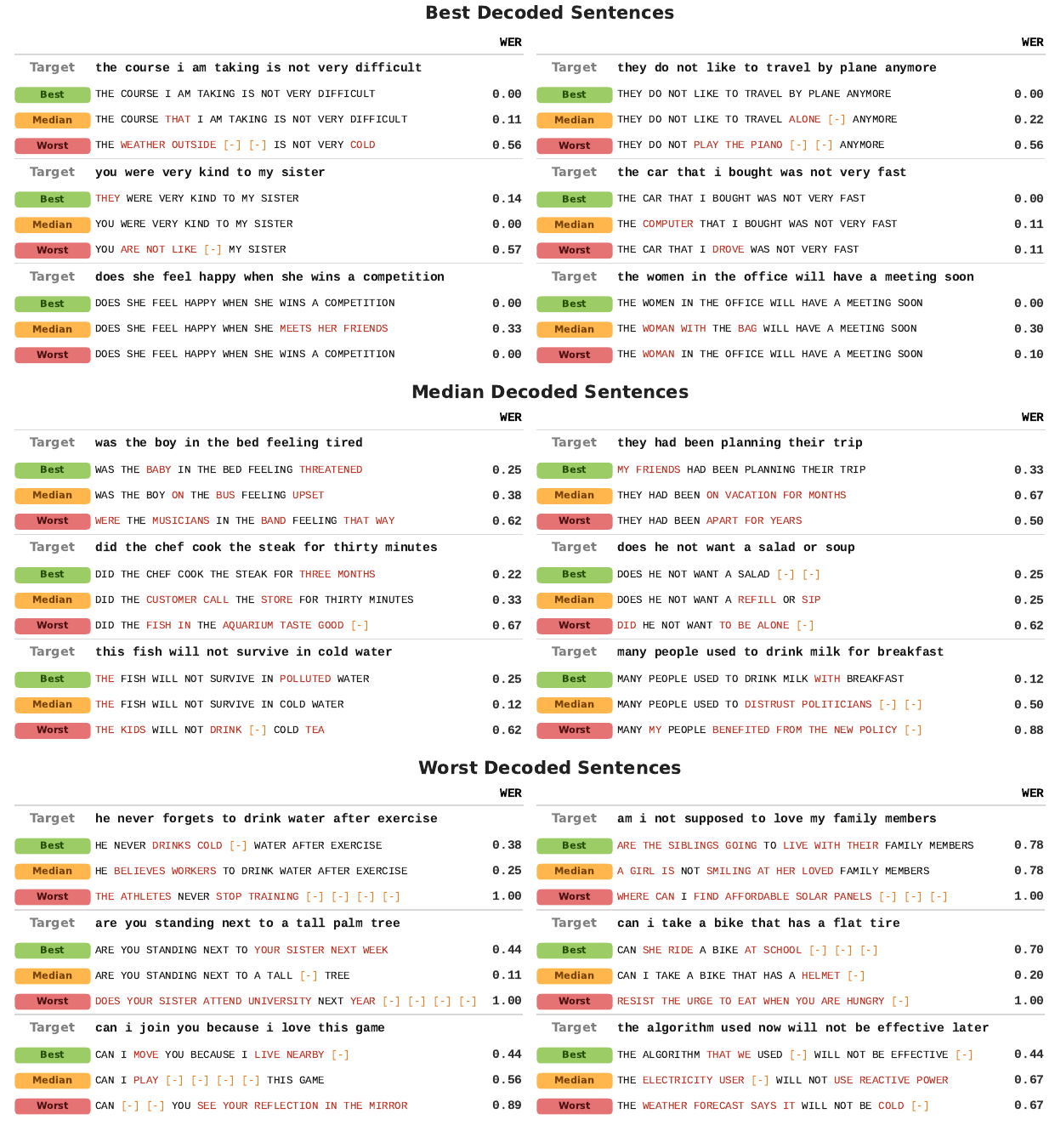}
    \caption{
        \textbf{Decoded sentence examples across difficulty bands.} \\
        18 sentences decoded by Brain2Qwerty~v2 for the \textbf{Best}, \textbf{Median}, and \textbf{Worst} subjects. Sentences are stratified by their mean Brain2Qwerty~v2 WER, averaged across all repetitions across all 9 subjects, into three terciles. Six sentences randomly drawn from each: \textbf{Best Decoded Sentences} (bottom tercile, top row), \textbf{Median Decoded Sentences} (middle tercile, middle row), and \textbf{Worst Decoded Sentences} (top tercile, bottom row). Each prediction is rendered with per-word colour coding: correct words in black, substitutions / insertions in red, and deletions marked [-] in orange. Per-row WER is shown on the right.
    }
    \label{fig:appendix-decoded-examples}
\end{figure}

Figure~\ref{fig:appendix-decoded-examples} shows 18 sentences and the corresponding Brain2Qwerty v2 decodings for the best, median, and worst subjects. Sentences are sorted by mean WER across all 9 subjects and split into three terciles — best, median, and worst decoded sentences — with 6 sentences sampled at random from each. 

On best decoded targets, the best subject is verbatim on 5 of 6 sentences and the median subject differs by at most one or two words. The errors that do appear are almost always meaning-preserving single-word substitutions --- ``travel alone'' for ``travel by plane'', ``computer'' for ``car'', or the insertion of ``that'' to maintain grammaticality. Even the worst subject can succeed verbatim on common templates (e.g., ``does she feel happy when she wins a competition'' is decoded perfectly by both the best and the worst subjects). On median decoded targets the substitutions become longer-range but the outputs remain grammatical and sometimes topic-adjacent. Examples: ``three months'' replaces ``thirty minutes'', ``they had been on vacation for months'' replaces ``they had been planning their trip''. On worst decoded targets the per-subject decodings diverge further still. For example: ``can i take a bike that has a flat tire'' yields ``a helmet'' (median) and ``can she ride a bike at school'' (best), while the worst subject produces an unrelated grammatical sentence. 

\clearpage
\subsection{Contribution of the Neuro Tokens}
\begin{figure}[h]
    \centering
    \includegraphics[width=\linewidth]{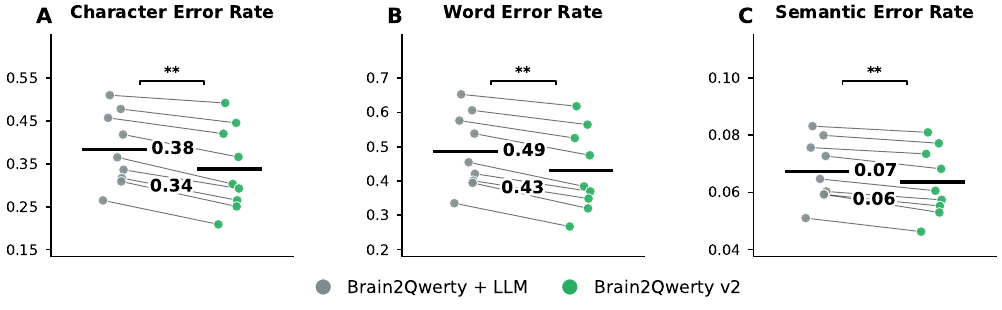}
    \caption{\textbf{Brain2Qwerty~v2 is a neuroLLM, not a corrector of CTC predictions.}\\
    \textbf{A--C.} Per-subject Character (\textbf{A}), Word (\textbf{B}), and Semantic (\textbf{C}) Error Rate for two configurations of our LLM-based decoder that share the same Qwen3-0.6B backbone. Brain2Qwerty~+\,LLM (grey) conditions the LLM only on the encoder's predictions; Brain2Qwerty~v2 (green) additionally conditions the LLM on the MEG-derived word embeddings (``Neuro Tokens''). Each pair of dots is one subject ($n=9$); thin black lines connect the same subject across the two configurations. Black horizontal bars are cross-subject means, printed beside each bar. Brackets: two-sided paired Wilcoxon signed-rank, ${**}\,p<0.01$.}
    \label{fig:neurotokens}
\end{figure}

A central question for an LLM-based decoder is whether the LLM is reading neural information or behaving as a corrector of the encoder's text output. To distinguish these two regimes, we ablate the Neuro Tokens, the MEG-derived word embeddings produced by our CTC tokenizer and word projector, and condition the LLM only on the encoder's predictions.
We refer to this ablation as Brain2Qwerty~+\,LLM. Brain2Qwerty~v2 improves on every metric: Character Error Rate ($0.34$ vs.\ $0.38$, Fig.~\ref{fig:neurotokens}\textbf{A}), Word Error Rate (\textbf{B}: $0.43$ vs.\ $0.49$), and Semantic Error Rate (\textbf{C}: $0.064$ vs.\ $0.067$). The gap is largest on Word Error Rate ($-5.6$~pts absolute), indicating that the Neuro Tokens carry word-level information that the CTC text alone is missing. These results establish that Brain2Qwerty~v2 is effectively modifying an LLM to decode directly from the neural signal.

\subsection{Encoder importance for downstream accuracy}
\begin{figure*}[h]
    \centering
    \includegraphics[width=\linewidth, trim={0 0.9cm 0 0}, clip]{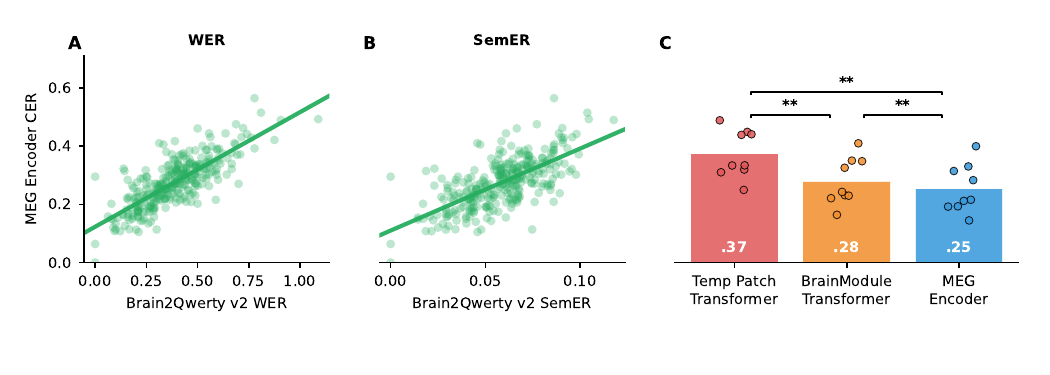}
    \caption{%
        \textbf{Encoder CER linearly predicts our model's performance, and architecture choices set the encoder CER.} \\ 
        Y-axis is shared by the three panels and represents the MEG~Encoder character error rate (CER) on the test set.
        \textbf{A.}~Per-sentence Brain2Qwerty~v2 word error rate (WER)
        versus MEG~Encoder CER, averaged across subjects per unique
        sentence. The green line is a linear regression.
        \textbf{B.}~Same as (A) with semantic error rate (SemER) on
        the x-axis.
        \textbf{C.}~Per-subject MEG~Encoder CER for three encoder architectures: Temporal Patch Transformer, BrainModule Transformer, and the MEG~Encoder used in this paper. All three pairwise comparisons are significant (paired Wilcoxon, two-sided).
    }
    \label{fig:encoder_importance}
\end{figure*}

To check whether the Encoder is the bottleneck of the pipeline, we ran two experiments. First, we correlated Brain2Qwerty~v2's per-sentence WER and SemER with the upstream encoder CER. Both downstream metrics scale linearly with Encoder CER, with Pearson coefficients of 0.78 for WER and 0.68 for SemER (Fig.~\ref{fig:encoder_importance}A,B): the better the keystroke predictions out of the encoder, the better the final decoded sentence. 

Because the upstream Encoder CER seems important, we compared three encoder architectures trained and tested on the same data: a Temporal Patch Transformer, inspired by~\cite{zhang2026crossspeciesneuralfoundationmodel} and first introduced in~\cite{feghhi2025time}; a BrainModule~+~Transformer (very similar to the Brain2Qwerty~v1 architecture) and first introduced in \cite{defossez2023decoding}; and the BrainModule~+~Conformer used in this paper. Both ablations result in significantly worse Encoder CER. Replacing the Conformer with a standard Transformer increases CER from 0.25 ($\pm0.03$) to 0.28 ($\pm 0.03$) ($p < 0.005$), demonstrating that the Conformer's interleaved convolutional and attention layers provide meaningful inductive bias for the local temporal structure of MEG signals. Further replacing the BrainModule with a Temporal Patch module increases CER to 0.37 ($\pm 0.03$) ($p<0.005$), the worst performance of all variants. The BrainModule thus provides architectural priors that are particularly important for the low signal-to-noise MEG inputs. Together, these results establish that both architectural components contribute jointly to the quality of the MEG Encoder output and as a consequence that of the final decoded sentences.

\end{document}